\documentclass[letterpaper]{article} 
\pdfoutput=1
\usepackage{aaai23}  
\usepackage{times}  
\usepackage{helvet}  
\usepackage{courier}  
\usepackage[hyphens]{url}  
\usepackage{graphicx} 

\usepackage{enumitem,kantlipsum}

\usepackage[utf8]{inputenc} 
\usepackage[T1]{fontenc}    
\usepackage{url}            
\usepackage{booktabs}       
\usepackage{amsfonts}       
\usepackage{nicefrac}       
\usepackage{microtype}      
\usepackage{xcolor}         
\usepackage{makecell}
\usepackage{wrapfig}
\usepackage{lipsum}
\usepackage{fancyhdr}
\usepackage[normalem]{ulem}
\usepackage{flushend}
\usepackage{subfigure}
\usepackage[ruled,noline,noend]{algorithm2e}

\usepackage{stfloats}
\usepackage{color}
\usepackage{graphicx}
\usepackage{amsmath}
\usepackage{setspace}
\usepackage{lipsum}
\usepackage{amsmath}
\usepackage{natbib}
\usepackage{float}

\usepackage{natbib}  
\usepackage{caption} 
\usepackage{newfloat}
\usepackage{listings}
\DeclareCaptionStyle{ruled}{labelfont=normalfont,labelsep=colon,strut=off} 
\floatstyle{ruled}
\newfloat{listing}{tb}{lst}{}
\floatname{listing}{Listing}
\title{Predictive Exit: Prediction of Fine-Grained Early Exits for Computation- and Energy-Efficient Inference}
\author{
    Xiangjie Li, Chenfei Lou, Yuchi Chen, Zhengping Zhu, Yingtao Shen, Yehan Ma, An Zou
}
\affiliations{
    \textsuperscript{}Shanghai Jiao Tong University \vspace{-4mm} \thanks{Published in AAAI-23. This research project is supported by NSFC 62202287, NSFC 62103268, and Shanghai Chenguang Program 21CGA11. An Zou is the corresponding author.}
    \\

}

\begin{document}

\maketitle

\begin{abstract}
By adding exiting layers to the deep learning networks, early exit can terminate the inference earlier with accurate results. 
However, the passive decision-making of whether to exit or continue the next layer has to go through every pre-placed exiting layer until it exits. In addition, it is hard to adjust the configurations of the computing platforms alongside the inference proceeds. By incorporating a low-cost prediction engine, we propose a Predictive Exit framework for computation- and energy-efficient deep learning applications. 
Predictive Exit can forecast where the network will exit (i.e., establish the number of remaining layers to finish the inference), which effectively reduces the network computation cost by exiting on time without running every pre-placed exiting layer. Moreover, according to the number of remaining layers, proper computing configurations (i.e., frequency and voltage) are selected to execute the network to further save energy. Extensive experimental results demonstrate that Predictive Exit achieves up to 96.2\% computation reduction and 72.9\% energy-saving compared with classic deep learning networks; and 12.8\% computation reduction and 37.6\% energy-saving compared with the early exit under state-of-the-art exiting strategies, given the same inference accuracy and latency. \vspace{-2mm}
\end{abstract}

\section{\uppercase\expandafter{\romannumeral1}. Introduction}
\label{sec:intro}
\begin{figure}
  \centering
  \includegraphics[trim={0cm 0cm 0cm 0cm},width=0.45\textwidth]{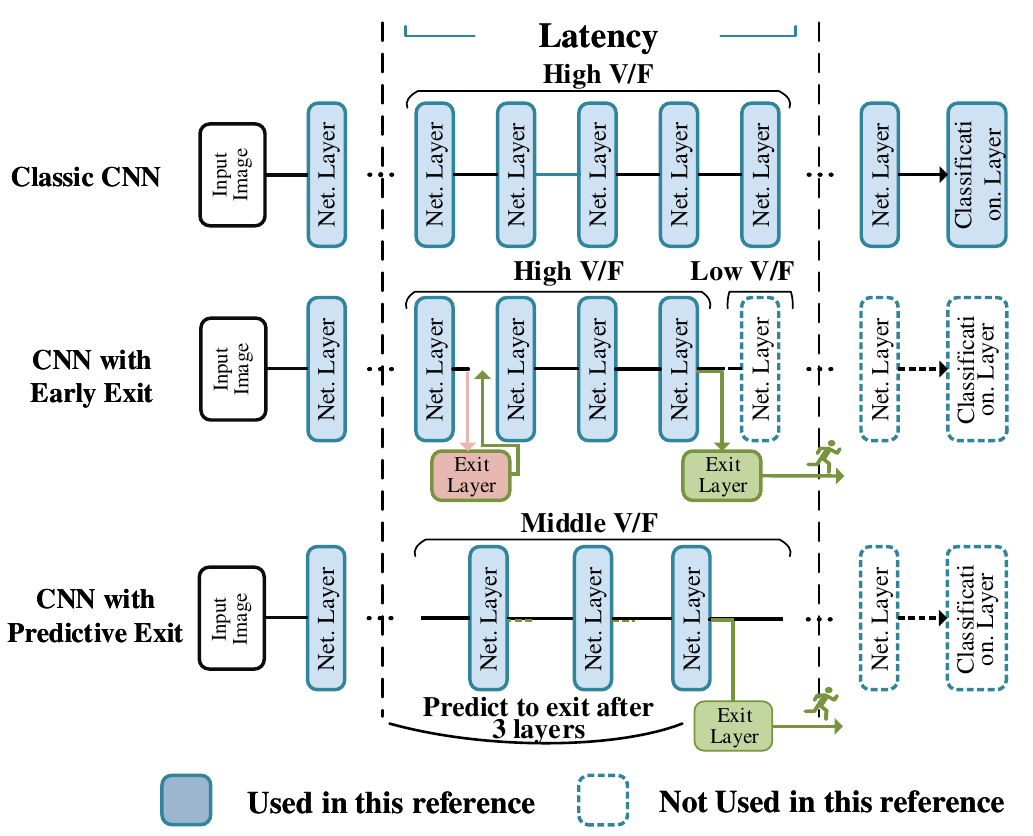}
  \caption{The architecture of classic deep learning network, early exit, and proposed Predictive Exit.}
  \label{fig:intro}
  \vspace{-5mm}
\end{figure}

Deep learning approaches, such as convolution neural networks (CNNs), have achieved tremendous success in versatile applications. However, deploying the deep learning models on resource-constrained systems is challenging because of its huge computation and energy cost. 
Diving into the performance of each inference case, researchers \citep{teerapittayanon2016branchynet,figurnov2017spatially,wang2018skipnet} found that the significant growth in model complexity is only helpful to classifying a handful of complicated inputs correctly, and they might become “wasteful” for simple inputs. Motivated by this observation, early exit includes additional side branch classifiers (exiting layers) to some of the network layers. As shown in Fig. \ref{fig:intro}, compared with the classic deep learning network in (a), the additional exiting layer in (b) allows
inference results for a large portion of test samples to exit the network early when samples have already been inferred with high confidence.

Despite being employed in burgeoning efforts to reduce computation cost and energy consumption in inference, early exits \citep{scardapane2020should} are not capable of addressing the challenges listed below.
First, there is a dilemma between fine-grained and coarse-grained placements of exiting points. Fine-grained exiting points lead to significant performance and energy overheads due to frequent execution of the exiting layers \citep{baccarelli2020optimized}. In contrast, coarse-grained exiting points may miss the opportunities to exit earlier. Second, the exiting layers have different topologies at different exiting points, which induce extra burdens to map the computation to computing hardware resources. Third, the energy-efficient computing platforms using early exits can only reduce the computing voltage and frequency after exiting. 
During the inference process, run-time computing configuration adjustment, such as dynamic voltage and frequency scaling (DVFS), can not be applied ahead of time for energy saving.

To provide an efficient early exit for resource-constrained computing platforms, we propose \emph{Predictive Exit}: prediction of fine-grain early exits in computation- and energy-efficient inference. 
To our best knowledge, Predictive Exit is the first work to forecast the exiting point and adjust the computing configuration (i.e. DVFS) along with the inference proceeds, which achieves significant computation and energy savings.
The contributions of this work are three-fold.

\begin{itemize}
\item A novel \emph{Predictive Exit framework} for deep learning networks is introduced, which is shown in Fig.~\ref{fig:intro} (c).
Fine-grained exiting layers sharing the same topology are \emph{potentially} placed after each network layer. A \emph{selective} execution of the exiting layers will capture the opportunities to exit on time and reduce computation and energy overheads. 

\item A low-cost \emph{prediction engine} is proposed to predict the point where the network will exit, execute the exiting layer selectively at run-time based on the prediction, and optionally adjust the computing configurations for energy saving during each inference based on DVFS. 

\item Extensive experiments with floating-point (FP32) and fixed-point quantization (INT8) operations are conducted on both server and embedded GPUs with energy measurements to demonstrate that 
Predictive Exit achieves significant computation cost reduction and energy saving with the state-of-the-art early exit strategies.\vspace{-2mm}
\end{itemize}

\section{\uppercase\expandafter{\romannumeral2}.Background and Related Work}
\label{sec:background_related}
\label{subsec:background}
\subsection{Latency, Power, and Energy}
In many applications, such as robots and self-driving cars, deep learning networks are released and executed periodically. The time interval between the release times of two consecutive inferences is called the inference period $T$. The actual execution time of each inference is called latency $l$, which rarely exceeds the period $T$. Given a computing platform and a neural network with fixed computation, the inference latency $l$ is inversely proportional to the operating frequency of the digital logic inside the computing platform, which is called the computing frequency $f$. Following the basic rules of semiconductors, a higher $f$ indicates faster computing but requires a linearly increased processor supply voltage $V$.

When a task is executed, the power consumption in computing, called \emph{active power} $P_{active}$, is the summation of dynamic, static, and constant power. When the computing platform is idle, the power consumption, also called \emph{idle power} $P_{idle}$, mainly comprises static and constant power.
The energy $E$ of a computing platform includes both active and idle power integrated with time,
\begin{tiny}
\begin{equation}
\begin{aligned}
    & E  =\int_{0}^{T_{active}}P_{active}dt + \int_{0}^{T_{idle}}P_{idle}dt\\
    &\!\! = \!\!\! \int_{0}^{T_{active}}\!\!(P_{Dynamic}\!\!+\!\!P_{Static}\!\!+\!\!P_{const})dt\!\!+\!\!\!\int_{0}^{T_{idle}}\!\!(P_{Static}\!\!+\!\!P_{const})dt
\end{aligned}
\label{eq:energy}
\end{equation}
\end{tiny}$T_{active}$ is the time spent on executing the task, and $T_{idle}$ is the time duration when the computing platform is idle, which is the time interval between finishing this inference and starting the next inference, denoted by $T_{idle}=T-T_{active}$. 
The \emph{dynamic power consumption} $P_{Dynamic}$ originates from the activity of logic gates inside a processor, which can be derived by
\begin{small}
\begin{equation}
    P_{Dynamic} = CV^{2}f,
\label{eq:dynamic}
\end{equation}
\end{small}where $C$ is the capacitance of switched logic gates.
The \emph{static power dissipation} $P_{Static}$ originates from transistor static current (including leakage and short circuit current) when the processor is powered on, which is described by
\begin{small}
\begin{equation}
    P_{Static} = VN_{tr}I_{static}.
\label{eq:static}
\end{equation}
\end{small}$N_{tr}$ is the number of logic gates, and $I_{\text{static}}$ is the normalized static current for each logic gate.
The \emph{constant power} $P_{const}$ is the power consumption by the auxiliary devices of a computing platform, like board fans and peripheral circuitry. 

\subsection{Early Exit with Dynamic Voltage Frequency Scaling}

DVFS is a power management technique in computer architecture whereby $f$ and $V$ of a microprocessor can be automatically adjusted on the fly depending on the actual needs to conserve power and energy cost of the computing platform. 
Given the same inference period $T$, combining early exit and DVFS could effectively reduce the cost of computation resources for a deep learning network.

Originally, the deep learning networks are executed by running all network layers at the default high frequency and voltage without early exit and DVFS.
We assume the inference latency (execution time) $l$ is equal to inference period $T$.
In this case, the energy consumption within $T$ is $E = \int_{0}^{T}(CV^2_{high}f_{high}+V_{high}N_{tr}I_{static}+P_{const})dt$.
As the deployment of early exists, 
network layers located ahead of the early exit run at the default high frequency and voltage. After the network exits at $T_{exit}$, computing platform can reduce the voltage and frequency to the lowest level by DVFS until time reaches $T$ \citep{tambe2021edgebert}. Therefore, the energy consumption can be described by $E = \int_{0}^{T_{exit}}(CV^2_{high}f_{high}+V_{high}N_{tr}I_{static}+P_{const})dt + \int_{T_{exit}}^{T}(V_{low}N_{tr}I_{static}+P_{const})dt$.

Since the processor voltage-frequency pair setting is cubic to $P_{Dynamic}$ but linear to $P_{Static}$ and latency $l$ according to Eqs.~\eqref{eq:dynamic} and ~\eqref{eq:static}, it is highly demanded to further reduce the cost of $P_{Dynamic}$. An intuition to achieve this is to adjust the voltage and frequency
to proper middle-level and run the network until it exits at time $T$. 
However, how to predict where the network will exit and adjust the voltage-frequency pair during the inference is not trivial.
In this work, we propose a solution to adjust the voltage and frequency to the proper ``middle-level''  at run-time based on the prediction of early exits. The energy cost is reduced by utilizing the inference period better.\vspace{-2mm}

\subsection{Related Work}
Early exits have attracted tons of attention as an important branch for dynamic inference in the past few years. Since it was studied in \citep{panda2016conditional, teerapittayanon2016branchynet}, many approaches have been proposed to improve the accuracy or reduce the computation cost of exit decisions. \cite{wang2019dynexit} offered a dynamic loss-weight adjustment early-exit strategy for ResNets together with the hardware-friendly architecture of early-exit branches. 
To determine the placement of exiting layers, \cite{kaya2019shallow} 
explored the distribution of computation cost; \cite{vanderlei2021class} 
classified the exit point for each inference object; 
\cite{panda2017energy} and \cite{baccarelli2020optimized}
calculated the benefit and additional computation and energy cost from adding this exiting layer. 
To further balance the accuracy, inference time, and energy tradeoffs, \cite{maciej2021zero} proposed a Zero Time Waste (ZTW) method approach by adding direct connections between exiting layers and combining previous outputs in an ensemble-like manner, and \cite{ghodrati2021frame} proposed an on-the-fly supervision training mechanism to reach a dynamic tradeoff between accuracy and power.


Together with the proceeding of theoretical models, early exits have started being deployed in hardware and IoT systems \citep{odema2021eexnas, samikwa2022adaptive}. \cite{li2020edge} and \cite{zeng2019boom} adopted device-edge synergy optimized by DNN partitioning and DNN right-sizing through the early exit for on-demand DNN collaborative inference. The strategy is feasible and effective for low-latency edge intelligence. \cite{xu2018efficient} presented a compressed CeNN framework optimized by five different incremental quantization methods and early exit. FPGA implementation shows that two optimizations achieve 7.8x and 8.3x speedup, respectively, while almost no performance loss. Meanwhile, \cite{laskaridis2021adaptive} and \cite{scardapane2020should} further provided a thorough overview of the current architecture, state-of-art methods, and future challenges of early-exit networks.

Although the multi-exit networks in these studies can lead to a reduction in inference time and computational cost, their efficiency lies in selecting a placement of exiting layers that keeps a good balance between cost-saving and inference accuracy. However, studies focused on choosing the best placement usually result in fixed places and thus neglect the possibility of exiting between exiting layers. Our work provides a predictive design that can dynamically change the placement of the exiting layer, which can further reduce inference time and computational cost.
\vspace{-2mm}
\label{subsec:related}
\begin{figure*}
  \centering
  \includegraphics[trim={0cm 0.5cm 0cm 0cm},width=0.78\textwidth]{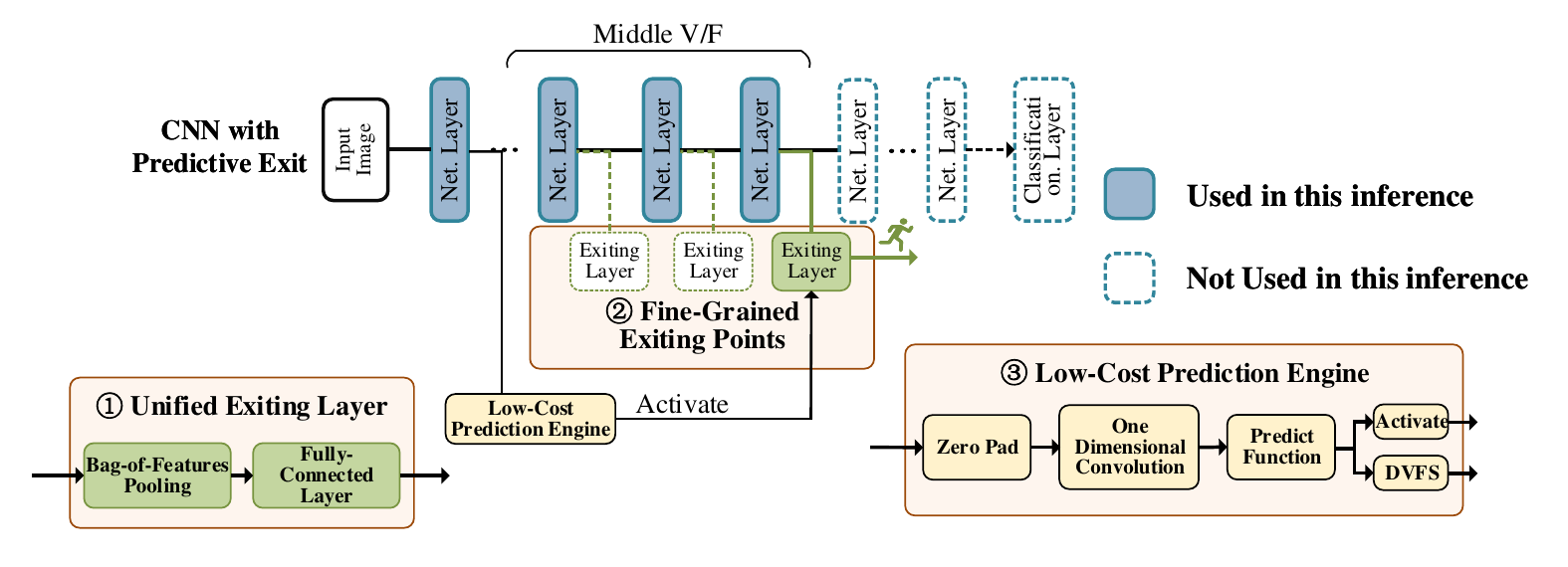}
  \caption{The framework of Predictive Exit.}
  \label{fig:frame}
  \vspace{-5mm}
\end{figure*}
\section{\uppercase\expandafter{\romannumeral3}. The Framework of Predictive Exit}
\label{sec:frame}
Targeting computation- and energy-efficient inference with early exits,
we propose a Predictive Exit framework for deep learning networks by taking the CNN as an example, which is the most popular deep learning algorithm.
The Predictive Exit combines the following schemes into one continuous spectrum, as shown in Fig. \ref{fig:frame}.

\textbf{Unified exiting layer:}
The neural networks run on the graphics processing units (GPU), tensor processing units (TPU), or application-specific integrated circuits (ASIC) on local or edge devices. To release the burden of executing versatile computation, we make the exiting layers used in the network share the same topology.

\textbf{Fine-grained exiting points:}
Exiting layers are potentially placed after each convolution layer to fully utilize the early exit opportunities. To avoid the computation overhead from running every exiting layer, only the layer at the predicted exiting point will be executed. 

\textbf{Low-cost prediction engine:}
The core of the framework is the low-cost prediction engine, which forecasts the possible exiting point and activates the early exiting layer. Based on the remaining computation workload before the expected exiting point, the proper computing frequency and voltage will be selected to run the inference and to further save energy. \vspace{-5mm}

\subsection{Unified Exiting Layer}
\vspace{-1mm}
In this work, the design of exiting layer is based on the existing work \citep{passalis2020efficient}. All the exiting layers share the same topology. It contains a Bag-of-Features (BoF) pooling layer and a Fully-Connected (FC) layer. Let $y_i$ be the intermediate results at the $i^{th}$ layer and $N_c$ be the number of object classes in the dataset. The BoF pooling layer functions as a feature aggregation to extract from $y_i$. In the BoF pooling, a set of feature vectors called codebook are used to describe $y_i$. The weight of each codebook is generated by measuring the similarity between the codebook and $y_i$. Since the size of codebook weight is usually larger than $N_c$, the FC layer further works as a classifier that adjusts the result of BoF pooling to $\mathbb{R}^{N_{C}}$, which estimates the final output of the exiting layer. Details can be found in \citep{passalis2020efficient}.
The function of the exiting layer is denoted as
\begin{small}
\begin{equation}
g_{\mathbf{W}_{i}}^{(i)}\left(y_i\right)=g_{\mathbf{W}_{i}}^{(i)}\left(f_{\mathbf{W}}(\mathbf{x}, i)\right) \in \mathbb{R}^{N_{C}},
\end{equation}
\end{small}where $f_{\mathbf{W}}(\mathbf{x}, i)$ stands for the intermediate result after the calculation of $1^{st}$ to $i^{th}$ layer.

After we've trained the early exiting layers, the average feature weight is first calculated according to Eq.~\eqref{eq:feature}, which are parameters for the exit decision.
\begin{small}
\begin{equation} 
\mu_{i}=\frac{1}{N} \frac{1}{N_{C}} \sum_{k=1}^{N_{c}} \sum_{j=1}^{N}\left[g_{\mathbf{W}_{i}}^{(i)}\left(y_i\right)\right]_{k},
\label{eq:feature}
\end{equation}
\end{small}where $N$ is the size of the training dataset. During the inference, at each early exiting layer, the weight ratio $\alpha_{W}$ is calculated as the maximum feature weight over the $\mu_{i}$ multiplied by a hyperparameter $\beta$ specified by the user. Once $\alpha_{W}$ in Eq.\eqref{eq:ratio} is larger than 1, the inference is terminated and the result at this early exiting layer is used as the final result.
\begin{small}
\begin{equation}
 \alpha_{W} = \frac{\left[g_{\mathbf{W}_{i}}^{(i)}\left(y_{j}\right)\right]_{k}}{\beta \mu_{i}}, k=\arg \max g_{\mathbf{W}_{i}}^{(i)}\left(y_{j}\right)
\label{eq:ratio}
\end{equation}
\end{small}The key idea behind such implementation is that the larger the maximum feature weight is, the more confident the classification is regarded. The parameter $\beta$ plays the role of striking a balance between the accuracy and acceleration. If a more accurate result is expected, $\beta$ should be higher and vice versa.\vspace{-2mm}

\subsection{Fine-Grained Exiting Points}
In the classic early exit design, a limited number of exiting layers are settled in a fixed location and distance, usually $\frac{1}{3}$ to $\frac{1}{4}$, of the neural networks. For example, in the inference model proposed in \citep{classicexample}, the two early exits are placed at roughly $\frac{1}{3}$ and $\frac{2}{3}$ of the network. If the first exiting layer fails, the inference must run through the following network layers until it reaches the next exiting layer to check whether it can exit. A more sensible way proposed by \cite{kaya2019shallow} is to settle the exiting layers based on the percentage of the total cost after a rough estimation of the computational costs of each layer. However, their ``fixed-place'' design ignores the possibility of exiting the layers between the two early exits. It is nontrivial to choose where and how to place early exits to satisfy both computation time and inference accuracy.

In this work, we proposed a fine-grained exiting points design, that is, placing potential exiting layers on every network layer since the starting layer $L_0$. $L_0$ is a hyperparameter specified by the user, working as an adjuster of computation time and inference accuracy. 
During the inference process, the trial of early exit begins on $L_0$. If the trial succeeds, the early exit is triggered. If fails, instead of running through every exiting layer from $L_0 + 1$, the position of the next trial is $L_1 = L_0 + \hbox{\sc Predict}\left(g_{\mathbf{W}_{L_0}}^{(L_0)}\left(y_{L_0}\right), \beta\right)$, where \hbox{\sc Predict} is the function of low-cost prediction engine to forecast the next location to exit, which will be introduced in the following subsection. Under the circumstances that early exit does not succeed on $L_{i}$, we will start another prediction based on the exiting layer result of $L_{i}$.
That is
\begin{small}
\begin{equation}
     \vspace{-1mm}
     L_{i+1} = L_i + \hbox{\sc Predict}\left(g_{\mathbf{W}_{L_i}}^{(L_i)}\left(y_{L_i}\right), \beta\right).
     \vspace{-1mm}
\end{equation}
\end{small}Therefore, the location expectation of exit $L_e$ can be expressed as 
\begin{small}
\begin{equation}
    L_e = L_0 + \sum_{j=1}^{\xi}p_{f_{j}} \times \hbox{\sc Predict}\left(g_{\mathbf{W}_{L_j}}^{(L_j)}\left(y_{L_j}\right), \beta\right),
\end{equation}
\end{small}where $L_0$ is the location of first trial, $\xi$ stands for the times of predictions, and $p_{f_{j}}$ is possibility of prediction failure.\vspace{-2mm}

\subsection{Low-Cost Prediction Engine}
\subsubsection{Prediction of Exiting Points}
The \hbox{\sc Predict} function starts to forecast the exiting point since the $L_0$ layer.
To fully use the information from the intermediate result during the inference process, the key function of \hbox{\sc Predict} is realized by one dimensional convolution on the intermediate result $y_{L_0}$ and the exiting layer result based on it $g_{\mathbf{W}_{L_0}}^{(L_0)}\left(y_{L_{0}}\right)$ at $L_0$, which is summarized in Algorithm 1.
To perform the convolution, we first extend the intermediate result $g_{\mathbf{W}_{L_0}}^{(L_0)}\left(y_{L_0}\right)$ from $\mathbb{R}^{N_c}$ to $\mathbb{R}^{K+N_c-1}$ by zero padding, which allow for more space for the filter to cover the intermediate result,
\begin{small}
\begin{equation}
\begin{aligned}
J_{L_0} & = \hbox{\sc ZeroPad}_{\left(K+N_c-1\right)}\left(g_{\mathbf{W}_{L_0}}^{(L_0)}\left(y_{L_0}\right)\right)\\
& = \begin{cases}g_{\mathbf{W}_{L_0}}^{(L_0)}\left(y_{L_0}\right)\left[i\right], & \frac{K-1}{2}\leq i \leq \frac{K-3}{2}+N_c\\ 0, & \text { otherwise }\end{cases}.
\label{eq:zero_pad}
\end{aligned}
\end{equation}
\end{small}Afterwards, a vector $h\in \mathbb{R}^{K}$ is generated (with all of 1 in our work) as the filter of the convolution. 
Based on $J_{L_0}$ and $h$, a new set of feature weight $G_{L_0}^{L_0+1}\in \mathbb{R}^{N_c}$ is generated through one dimensional convolution, which estimates the results of exiting layer placed at $L_0+1$,
\begin{small}
\begin{equation}
G_{L_0}^{L_0+1}[i]=\sum^{K-1}_{k = 0}h[k]\times J_{L_0}[i-\frac{k-1}{2}].
\label{eq:predict_g}
\end{equation}
\end{small}By recursively replacing the intermediate result $g_{\mathbf{W}_{L_0}}^{(L_0)}\left(y_{L_0}\right)$ in Eq. (\ref{eq:zero_pad}) with $G_{L_0}^{L_0+1}[i]$ and repeating the computation of Eq. (\ref{eq:zero_pad}) and Eq. (\ref{eq:predict_g}), the predicted results of exiting layer placed at $L_0+2$ can be obtained and noted as $G_{L_0+1}^{L_0+2}$. Following above steps, the predicted results of any exiting layer placed after $L_0$ can be calculated.

With the predicted results of exiting layer placed at any layer after $L_0$, we will have \hbox{\sc Predict} function described as follows:
\begin{small}
\begin{equation}
    \hbox{\sc Predict}\left(g_{\mathbf{W}_{L_0}}^{(L_0)}\left(y_{L_0}\right), \beta\right)  = \zeta ,
\end{equation}
\end{small}where $\zeta\in \mathbb{Z}$ represents the predicted exiting point after $L_0$, which is the smallest number that satisfies
\begin{small}
\begin{equation}
\frac{\left[G^{L_0+\zeta}_{L_0+\zeta-1}\right]_{k}}{\beta \mu_{L_0 + \zeta}} > 1, k=\arg \max G^{L_0+\zeta}_{L_0},
\label{x}
\end{equation}
\end{small}where $\frac{\left[G^{L_0+\zeta}_{L_0+\zeta-1}\right]_{k}}{\beta \mu_{L_0 + \zeta}}$ indicates the predicted exiting confidence at layer $L_0 + \zeta$.
Therefore, $L_{0}+\zeta$ is the predicted exiting point from \hbox{\sc Predict}. 
In case that $\zeta$ cannot be found, a hyperparameter $\tau$ is further introduced, that is $\zeta\in[1, \tau]$. $\tau$ should be no more than $L_{\text{total}} - L_0$ where $L_{\text{total}}$ stands for the number of layers of the model, meaning that prediction result beyond the last layer is forbidden. If no integer in $[1, \tau]$ satisfies Eq.  (\ref{x}), we assume that $\zeta = \tau$. 
The prediction engine is simple enough to avoid adding notable computation costs to the network. The time complexity of Algorithm 1 is $O(N_c \times K)$. Therefore, the computation cost of the prediction engine is $10^{-7}$ to $10^{-9}$ of the entire networks' computation cost. Meanwhile, hyperparameters are introduced in the prediction. $L_0$,  $\beta$, and $\tau$ can be tuned by advanced users, which can balance the prediction accuracy and computation cost for different application scenarios.
\begin{figure*}
\setlength{\abovecaptionskip}{-1mm}
\centering
\subfigure[Weight ratio $\alpha_{W}$]{
\label{fig:fig_schedulability_length_1} 
\includegraphics[width=0.4\textwidth]{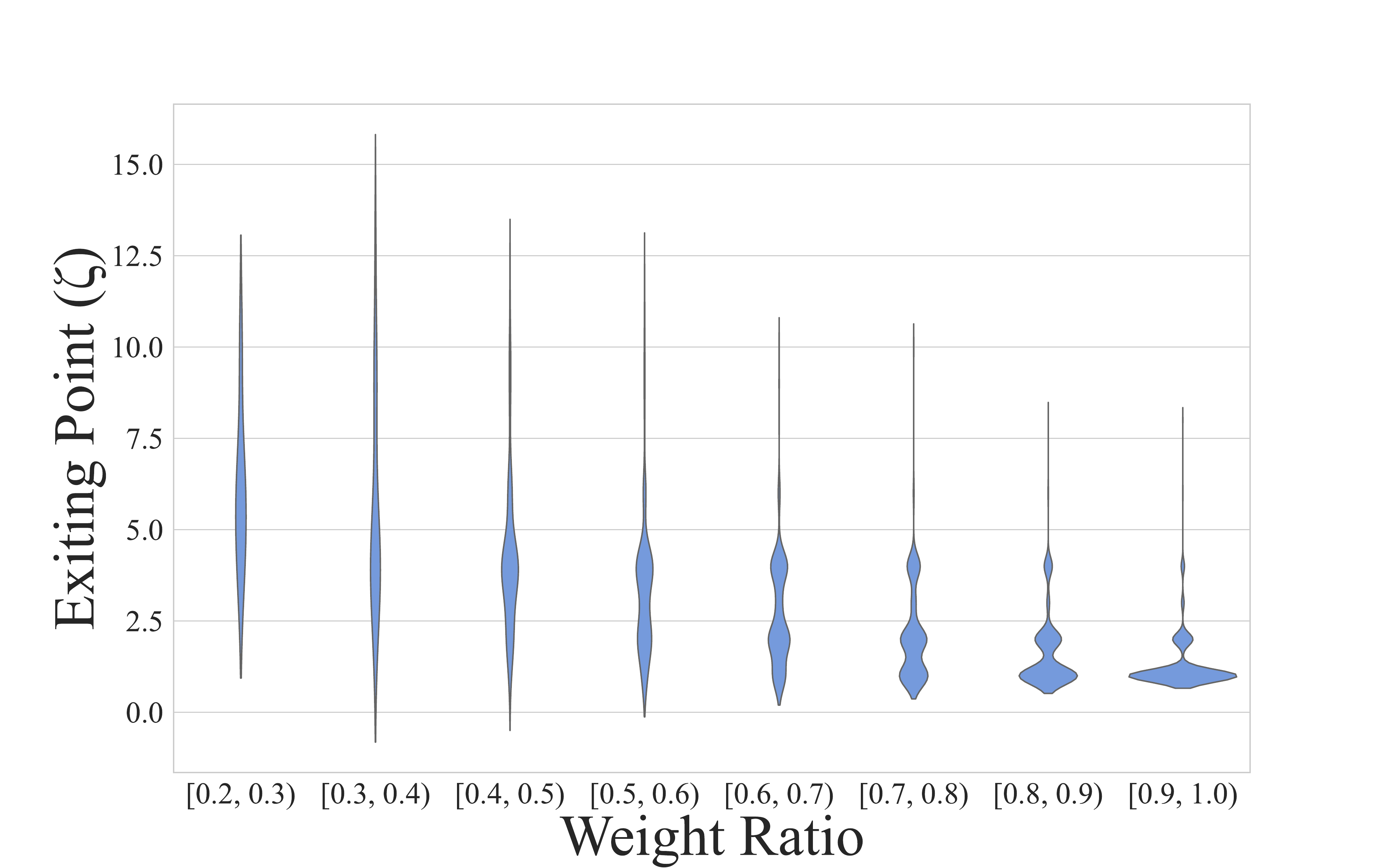}}
\subfigure[Cross entropy $\mathcal{J}(\mathrm{p}, \mathrm{q})$]{
\label{fig:fig_schedulability_length_2} 
\includegraphics[width=0.4\textwidth]{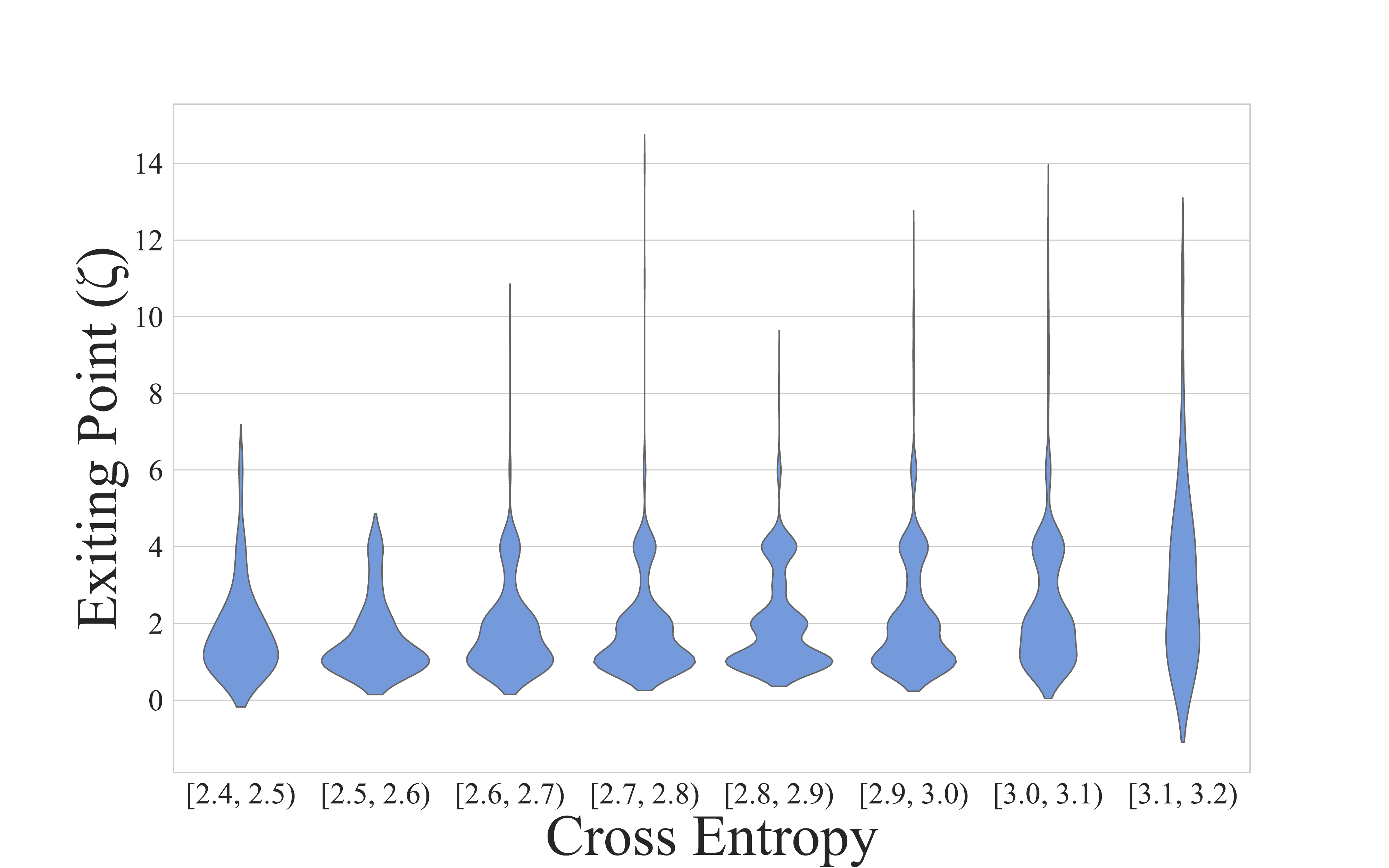}}
\caption{Exit prediction with weight ratio $\alpha_{W}$ and cross entropy $\mathcal{J}(\mathrm{p}, \mathrm{q})$.}
\label{fig:ratio_entropy} 
\vspace{-5mm}
\end{figure*}

\subsubsection{Discussion of Weight Ratio- and Cross Entropy-Based Prediction}
We discuss the prediction for early exits using 
$\alpha_{W}$ 
and cross entropy $\mathcal{J}(\mathrm{p}, \mathrm{q})$, which will be defined in Eq.~\eqref{eq:entropy}, since these two parameters are the indicators of confidence in early exits. More precisely, a higher weight ratio or a lower cross entropy indicates more confidence in exiting. As shown in Fig. \ref{fig:ratio_entropy},
we record the values of weight ratio and cross entropy ($x$ axis) at the prediction starting point $L_0$ and the actual exiting layer ($y$ axis) in the inference.
\begin{algorithm}
\begin{small}
\KwIn{Hyperparameter starting layer $L_{0}$, $\beta$ and $\tau$; \\ \quad\quad\quad Exiting layer result at $L_{0}$; $g_{\mathbf{W}_{L_0}}^{(L_0)}\left(y_{L_0}\right)$;}
\KwOut{Predicted exiting point: $L_{0}+\zeta$;}
\ArgSty{//The prediction procedure:\\}
\textbf{\scriptsize{1}} \For{$L_{0}+\zeta$ = $L_{0}+1$, ..., $L_{total}$}{
\:\:\ArgSty{\begin{small}//Zero padding from $\mathbb{R}^{N_c}$ into $\mathbb{R}^{K+N_c-1}$\\\end{small}}
\textbf{\scriptsize{2}} \If{$L_{0}+\zeta$ = $L_{0}+1$}
{
\begin{small}
$J_{L_0+\zeta}=\hbox{\sc ZeroPad}\left(g_{\mathbf{W}_{L_0}}^{(L_0)}\left(y_{L_0}\right)\right)$;\end{small}}
\:\:\:\:\Else{
$J_{L_0+\zeta}=\hbox{\sc ZeroPad}\left(G_{_{L_0+\zeta-1}}^{L_0+\zeta}\left(y_{L_0+\zeta-1}\right)\right)$;}
\:\:\ArgSty{\begin{small}//One dimensional convolution\\\end{small}}
\:\:\textbf{\scriptsize{3}} \begin{small}$G_{L_0+\zeta-1}^{L_0+\zeta}[i]=\sum^{K-1}_{k = 0}h[k]\times J_{L_0+\zeta}[i-\frac{k-1}{2}]$;\end{small}\\
\ArgSty{\begin{small}//Check prediction confidence\\\end{small}}
\textbf{\scriptsize{4}} \If{\begin{small}$\frac{\left[G^{L_0+\zeta}_{L_0+\zeta-1}\right]_{k}}{\beta \mu_{L_0 + \zeta}} > 1, k=\arg \max G^{L_0+\zeta}_{L_0}$\end{small}}
{\Return $\zeta$\;}
}
\:\:\textbf{\scriptsize{5}} $\zeta$ = $\tau$, where $\tau$ is a predefined hyperparameter.\\
\Return $\zeta$\;
\end{small}
\caption{\textbf{}: Low-cost Prediction Engine}
\label{Algorithm:schedulability_test}
\end{algorithm} 
The test case results of the Resnet-34 model and the SVHN dataset are presented in the figure.
However, on the weight ratios ranging from [0.2 0.6), the exiting points are distributed between the 1st and 12th layers. These spreading trends are also observed when the weight ratio lies in [0.6 0.8) and [0.8 1.0). Similarly, the exiting points are also widely distributed on the same cross entropy. Therefore, it is hard to predict the exiting point purely based on weight ratio and cross-entropy, even they are the indicators of confidence. \vspace{-2mm}

\subsection{DVFS for Predictive Exit}
Based on the predicted exiting point from the prediction engine, the voltage-frequency pair is adjusted to proper ``middle levels'' and run the network until it exits at time $T$. 
Given a network with $L_{total}$ layers and a prediction engine that starts the prediction at $L_{0}$ and predicts the exiting point is $L_{0}+\zeta$, the prediction engine will reduce the computing frequency (and its corresponding voltage) to 
\begin{small}
\begin{equation}
f_{middle} = \frac{L_{0}+\zeta-L_{0}}{L_{total}-L_{0}}f_{high},
\end{equation}
\end{small}where $f_{middle}$ is the lowest frequency that can finish the inference by time $T$ and $f_{high}$ is the default high frequency of the computing platform.
Therefore, the energy consumption used in this inference would be

\begin{small}
\begin{equation}
\begin{aligned}
E = \int_{0}^{T_{L_{0}}}(CV^2_{high}f_{high}\!\!+\!\!V_{high}N_{tr}I_{static}\!\!+\!\!P_{const})dt \\
+ \int_{T_{L_{0}}}^{T}(CV^2_{middle}f_{middle}\!\!+\!\!V_{middle}N_{tr}I_{static}\!\!+\!\!P_{const})dt,
\end{aligned}
\end{equation}
\end{small}where $T_{L_{0}}$ is the time of finishing the prediction starting layer $L_{0}$ and its previous network layers. As the selection of computing frequency is based on the predicted remaining workloads in the inference, the prediction engine can make the inference finish by time $T$ based on correct predictions. Since the processor voltage and frequency are linear scales to the computing performance (inversely proportional to inference latency) but the cubic scale and linear scale to dynamic power and static power, the predictive exit will effectively reduce energy consumption. \vspace{-2mm}
\section{\uppercase\expandafter{\romannumeral4}. Training the Network with Predictive Exit}
The learning objective of Predictive Exit is to keep the network inference accuracy given the predictive exit functions. Therefore, the objective function is the cross entropy loss
\begin{small}
\begin{equation}
\label{eq:entropy}
\mathcal{J}(\mathrm{p}, \mathrm{q})=-\sum_{i=1}^{N_{C}}\mathrm{p}(i) \log \left(\mathrm{q}(i)\right),
\end{equation}
\end{small}where $\mathrm{p}(i)$ and $\mathrm{q}(i)$ are the true class distribution and the predicted class distribution for each object.

Our model is trained through batch gradient descent. The data are fed into the model to optimize the parameter. Let $N_{\text{train}}$ be the size of the training set $\mathcal{X}$, and a target vector $r\in \mathbb{R}^{N_c}$ be the correct result.
At first, the model is trained without any exiting layer by the following equation
\begin{small}
\begin{equation}
W'=W-\eta \cdot \nabla_{W} \mathcal{J}\left(f_w\left(\mathcal{X}, L_{total}\right), r\right),
\end{equation}
\end{small}where $\eta$ is the learning rate. In the circumstance that the accuracy fails to meet the requirement, $\eta$ can be adjusted. 
Afterward, with $W$ of the original model fixed and armed with exiting layers, the  model is trained again to optimize the parameters in the BoF pooling and FC layer,
\begin{small}
\begin{equation}
W'=W-\eta \cdot \nabla_{W} \mathcal{J}\left(g_{\mathbf{W}_{i}}^{\left(i\right)}\left(y_{i}\right), r\right), i \in \left[1, ..., L_{\text{total}} \right].
\end{equation}
\end{small}
Although the training procedure requires two steps, the advantage brought by the inference process far outweighs this disadvantage, especially when only the inference is deployed on the resource-constrained platforms. 
\section{\uppercase\expandafter{\romannumeral5}. Evaluation}
\label{sec:evaluation}
\vspace{-2mm}
\begin{table*}[t]
\begin{scriptsize}
\centering
\setlength{\abovecaptionskip}{2mm}
\setlength{\belowcaptionskip}{1mm}
\caption{Inference accuracy and computation cost in the VGG-19 and ResNet-34 network}
\centering
\begin{tabular}{ccccc|cccc}
\toprule
\multicolumn{1}{c}{\textbf{Model}} & \multicolumn{4}{c}{\textbf{VGG-19}} & \multicolumn{4}{c}{\textbf{ResNet-34}}\\
\makecell{\textbf{Approach}} & \textbf{Classic CNN} & \textbf{Hierarchical} & \textbf{Placement} & \textbf{Predictive Exit} & \textbf{Classic CNN} & \textbf{Hierarchical} & \textbf{Placement} & \textbf{Predictive Exit}\\
\midrule
\makecell{FP32 CIFAR-10} & 89\% | 100\% & 88\% | 59.8\% & 88\% | 46.9\% & \textbf{88\% | 46.2\%} & 90\% | 100\% & 89\% | 28.7\% & 89\% | 12.1\% & \textbf{89\% | 7.7\%}\\

\makecell{FP32 CIFAR-100} & 64\% | 100\% & 62\% | 86.6\% & 62\% | 80.6\% & \textbf{62\% | 76.2\%} & 66\% | 100\% & 63\% | 36.6\% & 63\% | 21.2\% & \textbf{63\% | 16.1\%}\\

\makecell{FP32 SVHN} & 89\% | 100\% & 87\% | 55.6\%  & 87\% | 56.5\% & \textbf{87\% | 46.8\%} & 91\% | 100\% & 89\% | 6.9\% & 89\% | 5.4\% & \textbf{89\% | 3.8\%}\\

\makecell{FP32 STL10} & 75\% | 100\% & 74\% | 67.6\%  & 74\% | 64.2\% & \textbf{74\% | 51.5\%} & 76\% | 100\% & 73\% | 61.8\% & 73\% | 17.8\% & \textbf{73\% | 5.7\%}\\

\makecell{Int8 CIFAR-10} & 88\% | 100\% & 87\% | 62.3\% & 87\% | 46.4\% & \textbf{87\% | 46.1\%} & 85\% | 100\% & 82\% | 29.6\% & 82\% | 14.9\% & \textbf{82\% | 8.4\%}\\

\makecell{Int8 CIFAR-100} & 64\% | 100\% & 62\% | 95.0\% & 62\% | 79.5\% & \textbf{62\% | 76.4\%} & 62\% | 100\% & 60\% | 44.4\% & 60\% | 22.3\% & \textbf{60\% | 18.5\%}\\

\makecell{Int8 SVHN} & 89\% | 100\% & 86\% | 57.2\%  & 86\% | 47.4\% & \textbf{86\% | 46.7\%} & 88\% | 100\% & 85\% | 7.1\% & 85\% | 5.6\% & \textbf{85\% | 4.0\%}\\

\makecell{Int8 STL10} & 74\% | 100\% & 74\% | 66.9\%  & 74\% | 64.3\% & \textbf{74\% | 51.5\%} & 71\% | 100\% & 70\% | 78.1\% & 70\% | 17.8\% & \textbf{70\% | 5.8\%}\\
\bottomrule
\end{tabular}
\label{tab:inference_vgg}
\end{scriptsize}
\vspace{-4mm}
\end{table*}

\begin{figure*}
\setlength{\abovecaptionskip}{-2mm}
\centering
\subfigure[VGG-19 network]{
\label{fig:vgg_accuracy} 
\includegraphics[width=0.85\textwidth]{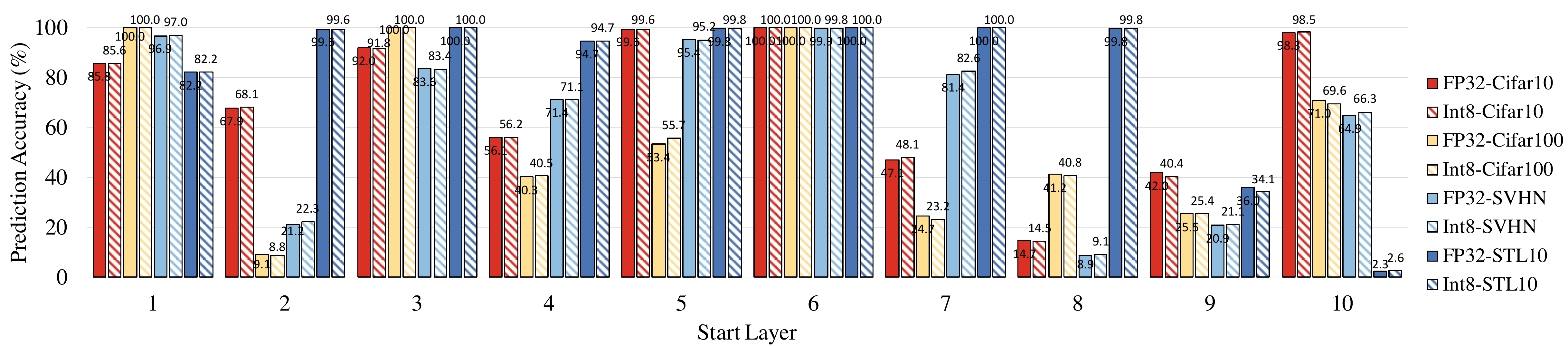}
\vspace{-4mm}
}
\subfigure[ResNet-34 network]{
\label{fig:resnet_accuracy} 
\includegraphics[width=0.85\textwidth]{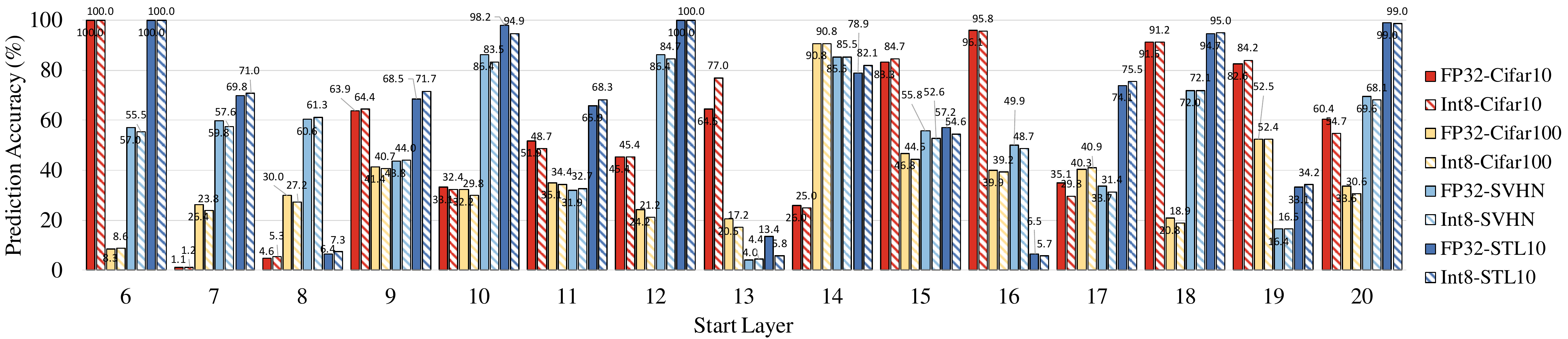}}
\caption{Prediction accuracy with hyperparameter $L_0$.}
\label{fig:L0_accuracy} 
\vspace{-5mm}
\end{figure*}

\subsection{Experimental Setup}
We evaluate the Predictive Exit using \textbf{VGG-19} and \textbf{ResNets-34} as the backbone models on the commonly used CIFAR-10, CIFAR-100 \citep{cifar10}, SVHN \citep{svhn}, and STL10 datasets \citep{stl10}. For the consideration of low computation and energy applications, both floating-point (FP32) and fixed-point quantization (Int8) operations are tested. 
The training follows the procedure in Section 4. For the classic network and exiting layer, we set the learning rates as 0.00025/0.001 and train them with 100 iterations.
We compare the proposed \textbf{Predictive Exit} with different early exit approaches:
\begin{itemize}
\item\textbf{Classic CNN} adopts VGG-19 \citep{simonyan2014very} or ResNets-34 \citep{he2016deep}; 
\item\textbf{Hierarchical} Early Exits, proposed in \citep{passalis2020efficient}, which is specifically designed for CNNs by placing exits at roughly $\frac{1}{4}$, $\frac{1}{2}$, $\frac{3}{4}$ of the network; 
\item\textbf{Placement} Early Exit, proposed in \citep{panda2017energy} and improved in \citep{baccarelli2020optimized} with tunable hyperparameters, which identifies the best place to attach an early exit by considering the benefit and extra computation and energy cost.
\end{itemize}
The computation and energy cost are evaluated on both server GPU Quadro GV100 and embedded GPU Jetson TX2 with both offline and online measured power consumption. The overhead from executing the prediction engine and adjusting the processor voltage and frequency are included in the tests. \vspace{-5mm}
\subsection{Inference Accuracy and Computation Cost}
The inference accuracy and computation cost (the number of floating-point or integer operations normalized by that of classic CNN) of different inference approaches and datasets are shown in Table \ref{tab:inference_vgg}.
On the VGG-19 model, Predictive Exit achieves the same inference accuracy as other early exit approaches (which is 1\% - 3\% lower than Classic CNN). Because Predictive Exit will continue for the next prediction instead of forcing the inference to exit even if the prediction is wrong, no extra inference accuracy loss is brought. 
Compared with Hierarchical and Placement, 
Predictive Exit further reduces up to 12.8\% of computation cost by leveraging the opportunities to exit earlier and avoiding frequent execution of the exiting layers. On the ResNet model, compared with Hierarchical and Placement, Predictive Exit reduces up to 12.1\% of computation cost without extra accuracy loss.

For the training process, compared with Classic CNN, which requires only the training of original network layers, Hierarchical structure demands additional training of the exiting layers placed in specified positions of the network. While both Placement and Predictive design need to train all exiting layers that settle after each layer of the network, Placement further needs one more step to determine placement indexes by an exhaustive search for the most profitable exiting layers. \vspace{-4mm}

\begin{table*}
\centering
\begin{scriptsize}
\setlength\tabcolsep{1.5pt}
\setlength{\abovecaptionskip}{2mm}
\setlength{\belowcaptionskip}{1mm}
	\caption{Measured power consumption of the NVIDIA Quadro GV100  GPU system}
	\centering
	\begin{tabular}{ccccccccccccccccccc}
		\toprule
\textbf{Frequency (GHz)} & \textbf{0.60} & \textbf{0.65} & \textbf{0.70} & \textbf{0.75} & \textbf{0.80} & \textbf{0.85} & \textbf{0.90} & \textbf{0.95} & \textbf{1.00} & \textbf{1.05} & \textbf{1.10} & \textbf{1.15} & \textbf{1.20} & \textbf{1.25} & \textbf{1.30} & \textbf{1.35} & \textbf{1.40} & \textbf{1.45} \\
\midrule
\makecell{Active Power (W)} & 59.2 & 67.4 & 73.5 & 81.1 & 85.3 & 90.4 & 97.5 & 104.9 & 112.8 & 119.5 & 130.1 & 139.5 & 148.9 & 161.1 & 170.2 & 180.6 & 199.1 & 218.5\\
\makecell{Idle Power (W)} & 35.1 & 35.9 & 37.0 & 37.8 & 38.9 & 39.8 & 41.1 & 41.3 & 43.8 & 44.2 & 45.0 & 45.8 & 46.5 & 47.8 & 49.3 & 50.5 & 52.3 & 55.1\\
\bottomrule
\end{tabular}
\label{tab:power}
\end{scriptsize}
\vspace{-4mm}
\end{table*}

\begin{tiny}
\begin{table*}
\centering
\begin{scriptsize}
\setlength{\belowcaptionskip}{1mm}
\setlength{\abovecaptionskip}{2mm}
\setlength\tabcolsep{1.5pt}
	\caption{Normalized energy consumption}
	\centering
	\begin{tabular}{ccccc|cccc}
\toprule
\multicolumn{1}{c}{\textbf{Model}} & \multicolumn{4}{c}{\textbf{VGG-19}} & \multicolumn{4}{c}{\textbf{ResNet-34}}\\
\textbf{Approach} & \textbf{Classic CNN} & \textbf{Hierarchical} & \textbf{Placement} & \textbf{Predictive Exit} & \textbf{Classic CNN} & \textbf{Hierarchical} & \textbf{Placement} & \textbf{Predictive Exit}\\
\midrule
\makecell{FP32 CIFAR-10} & 100\% & 62.2\% & 47.9\%  & \textbf{27.1\%} & 100\% & 58.6\% & 59.3\% & \textbf{30.3\%}\\
\makecell{FP32 CIFAR-100} & 100\% & 85.1\% & 72.8\% & \textbf{44.1\%} & 100\% & 69.9\% & 67.3\% & \textbf{39.9\%}\\
\makecell{FP32 SVHN} & 100\% & 58.9\% & 54.0\% & \textbf{27.1\%} & 100\% & 46.8\% & 49.1\% & \textbf{27.1\%}\\
\makecell{FP32 STL10} & 100\% & 65.7\% & 64.7\% & \textbf{27.1\%} & 100\% & 73.6\% & 64.5\% & \textbf{27.2\%}\\
\makecell{Int8 CIFAR-10} & 100\% & 76.8\% & 47.0\% & \textbf{27.1\%} & 100\% & 58.8\% & 60.7\% & \textbf{30.6\%}\\
\makecell{Int8 CIFAR-100} & 100\% & 85.2\% & 73.1\% & \textbf{45.1\%} & 100\% & 75.0\% & 68.1\% & \textbf{41.9\%}\\
\makecell{Int8 SVHN} & 100\% & 54.0\% & 47.7\% & \textbf{27.1\%} & 100\% & 46.9\% & 49.1\% & \textbf{27.1\%}\\
\makecell{Int8 STL10} & 100\% & 66.0\% & 64.7\% & \textbf{27.1\%} & 100\% & 85.3\% & 64.5\% & \textbf{27.2\%}\\
\bottomrule
\end{tabular}
\label{tab:energy}
\end{scriptsize}
\vspace{-5mm}
\end{table*}
\end{tiny}

\subsection{Where to Start Prediction: Hyperparameter $L_0$}
\label{sec:predict_accuracy}
The prediction engine will start the exiting prediction since the starting layer $L_0$. For each model and dataset, choosing $L_0$ should be done before the deep learning model is deployed as it directly determines Predictive Exit's performance. This section quantitatively compares the prediction accuracy of different $L_0$ during the training process. 
To let the prediction engine cover a wide range of network layers, $L_0$ should be the first few layers in the networks. Therefore, in the VGG-19 network, we test $L_0$ placed at the 1st-10th layers, and in the ResNet-34 network, we test $L_0$ placed at the 6th-20th layers. 
Fig. \ref{fig:vgg_accuracy} and Fig. \ref{fig:resnet_accuracy} present the prediction accuracy under FP32 and INT8 operations across different datasets. The prediction accuracy indicates the percentage of successful exiting at the first predicted exiting point.
In the VGG-19 network, if $L_0$ is the first network layer, the prediction can achieve over 81\% accuracy across all tested datasets and operations. Most notably, if $L_0$ is the 6th network layer, the prediction can achieve over 99.8\% accuracy. Therefore, $L_0$ for VGG-19 network and test datasets will be the 6th network layer. Similarly, for the ResNet-34 network, the desired $L_0$ will be the 6th, 14th, 10th, and 6th layers for datasets CIFAR-10, CIFAR-100, SVHN, and STL 10, respectively. \vspace{-2mm}
\subsection{Energy Benefit by DVFS}
To illustrate the potential energy benefit of Predictive Exit, we first calculate the energy consumption with offline measured active and idle power of NVIDIA Quadro GV100 GPU at different frequency-voltage pairs \citep{kandiah2021accelwattch} (shown in Table \ref{tab:power}). In a network with $L_{total}$ layers, when the Predictive Exit predicts the network will exit at $L_0+\zeta$ layers, the processor will select the lowest candidate frequency in Table \ref{tab:power} that is higher than or equal to $\frac{L_{0}+\zeta-L_{0}}{L_{total}-L_{0}}*1.45GHz$ to execute this network. We count the inference workload (including the exiting layer) under each frequency-voltage pair. Based on the power consumption in each frequency-voltage pair, we compare the energy consumption and normalize it to the energy consumption used by the Classic CNN model in Table \ref{tab:energy}. Unsurprisingly, early exit achieves tremendous energy savings compared with Classic CNN models. Predictive Exit further reduces the energy consumption (compared with the best cases of Hierarchical and Placement) by 19.9\% to 37.6\% on the VGG-19 network and 19.7\% to 37.3\% on the ResNet-34 network, by predicting the exiting and adjusting the computation configuration along with the inference process. Meanwhile, the proposed Predictive Exit is also tested on the embedded Nvidia Jetson TX2, whose power consumption is measured at run time with Tektronix mdo32 oscilloscope and TCP2020 current probe. Since its limited memory is incapable of loading \textbf{ResNets-34} models with early exits, experiments using \textbf{VGG-19} are tested for power measurement and its results are summarized in Table \ref{tab:tx2_energy}. The Predictive Exit reduces up to 45.7\% and 32.2\% of energy compared with Hierarchical and Placement early exits. \vspace{-2mm}

\begin{tiny}
\begin{table}
\centering
\begin{scriptsize}
\setlength{\belowcaptionskip}{1mm}
\setlength{\abovecaptionskip}{2mm}
\setlength\tabcolsep{1.5pt}
	\caption{Energy consumption on Jetson TX2 (Joule)}
	\centering
	\begin{tabular}{ccccccccc}
\toprule
\textbf{Approach} & \textbf{Classic CNN} & \textbf{Hierarchical} & \textbf{Placement} & \textbf{Predictive Exit}\\
\midrule
\makecell{CIFAR-10} & 1195.6 & 451.9 & 500.2  & \textbf{384.8} \\
\makecell{CIFAR-100} & 1198.2 & 914.8 & 732.2  & \textbf{496.6}\\
\makecell{SVHN} & 1209.4 & 873.9 & 521.6  & \textbf{492.2}\\
\bottomrule
\end{tabular}
\label{tab:tx2_energy}
\end{scriptsize}
\vspace{-5mm}
\end{table}
\end{tiny}

\section{\uppercase\expandafter{\romannumeral6}. Conclusion}
\label{sec:conclusion}
The proposed Predictive Exit can accurately predict where the network will exit as a computation- and energy-efficient inference technique. By activating the exiting layer at the expected exiting point, the Predictive Exit reduces the network computation costs by exiting on time without running every pre-placed exiting layer. The Predictive Exit significantly reduces the energy consumption used in inference by selecting proper computing configurations in the inference process. The current design and evaluation are based on CNN topology. Since our method is a plug-in for existing networks without model modification, the Predictive Exit could be applied to general learning networks such as Transformer and MLP, which will be explored in our future work.
\bibliography{main}
\end{document}